\documentclass[runningheads]{llncs}
\UseRawInputEncoding
\usepackage{multirow}
\usepackage{amssymb}
\usepackage{mathtools}

\begin{document}
\title{Evidence-aware multi-modal data fusion and its application to total knee replacement prediction}
\titlerunning{Evidence-aware multi-modal data fusion for TKR prediction}

\author{Xinwen Liu\inst{1} \and Jing Wang\inst{2} \and S. Kevin Zhou\inst{3,4} \and Craig Engstrom \inst{5} \and \\ Shekhar S. Chandra \inst{1}}
\authorrunning{X.Liu, et al.}

\institute{School of Information Technology and Electrical Engineering, \\The University of Queensland, Brisbane, Australia.\\
\and The Commonwealth Scientific and Industrial Research Organisation, \\ Canberra, Australia. \\
\and Center for Medical Imaging, Robotics, Analytic Computing \& Learning (MIRACLE), School of Biomedical Engineering \& Suzhou Institute for Advanced Research, University of Science and Technology of China, Suzhou, China.\\
\and Key Laboratory of Intelligent Information Processing of Chinese Academy of Sciences (CAS), Institute of Computing Technology, CAS, Beijing, China.\\
\and School of Human Movement and Nutrition Sciences, \\ The University of Queensland, Brisbane, Australia.}
\maketitle 

\begin{abstract}
Deep neural networks have been widely studied for predicting a medical condition, such as total knee replacement (TKR). It has shown that data of different modalities, such as imaging data, clinical variables and demographic information, provide complementary information and thus can improve the prediction accuracy together. However, the data sources of various modalities may not always be of high quality, and each modality may have only partial information of medical condition. Thus, predictions from different modalities can be opposite, and the final prediction may fail in the presence of such a conflict. Therefore, it is important to consider the reliability of each source data and the prediction output when making a final decision. In this paper, we propose an evidence-aware multi-modal data fusion framework based on the Dempster-Shafer theory (DST). The backbone models contain an image branch, a non-image branch and a fusion branch. For each branch, there is an evidence network that takes the extracted features as input and outputs an evidence score, which is designed to represent the reliability of the output from the current branch. The output probabilities along with the evidence scores from multiple branches are combined with the Dempster's combination rule to make a final prediction. Experimental results on the public OA initiative (OAI) dataset for the TKR prediction task show the superiority of the proposed fusion strategy on various backbone models. 
\keywords{Dempster-Shafer theory \and Medical condition prediction \and Multi-modal data fusion}
\end{abstract}

\section{Introduction}
Deep learning (DL) has been increasingly investigated for medical condition and disease prediction at the early stage. DL recommendations for early interventions can potentially lead to a better treatment outcome, reduce healthcare cost, and save the life of patients. Deep neural networks have been leveraged to support clinical decision-making by analyzing a large amount of data consisting of health records, medical images, and clinical information \cite{guan2021mri,hoang2022clinically,kline2022multimodal,rana2020multi,venugopalan2021multimodal,zheng2022multi}. Data from multiple modalities normally contain complementary information, and the effective combination of multi-modal information can improve the prediction and diagnosis accuracy \cite{bayoudh2021survey,cheng2021brain,guan2022deep,li2020self,liu2021incomplete,mallya2022deep}. 

Multi-modality learning has shown to be effective in many predictive tasks such as Alzheimer's disease (AD) progression prediction \cite{guan2021mri,huang2022multi,lee2019predicting,zhang2011multi}, knee osteoarthritis (OA) trajectory forecasting \cite{hoang2022clinically,joseph2022machine,karim2021deepkneeexplainer,nguyen2021climat,tiulpin2019multimodal}, COVID-19 mortality risk prediction \cite{rahman2022bio} and peripapillary atrophy forecasting \cite{li2021causal}. For example, Knee OA is the most common musculoskeletal (MSK) disorder. When diagnosed early, knee OA can be managed through non-invasive treatment options; otherwise, the only option is the total knee replacement (TKR) surgery. The successful prediction of TKR can assist with the optimal choice of treatment options. Authors in~\cite{tolpadi2020deep} combine the use of magnetic resonance (MR) images and clinical data for an accurate TKR prediction.

However, the prediction outcomes from each modality can be different, because the quality of the source data may not always be good and the information inherent in one modality only provides partial support for decision-making~\cite{li2022confidence}. The final decision can be catastrophically wrong without a proper consideration of such a  variance in various sources, although one source could have provided a right prediction. For example, when analyzing TKR prediction experiments with existing methods~\cite{tolpadi2020deep}, we find that \textit{there are nearly half failure cases could have been classified correctly with a single modality prediction}. 

To alleviate this issue and improve the prediction confidence and accuracy, we propose an evidence-aware multi-modal data fusion framework for the  medical condition prediction based on Dempster-Shafer theory (DST). Specifically, an evidence network is placed in parallel to the backbone prediction branch. The evidence networks take the extracted features as input and return evidence scores to quantify the confidence of the output. The evidence scores and the decision outputs of all sources are combined based on Dempster's combination rule for a final prediction. The induced-results consider the probabilities and the evidence from all the sources, making a more reliable final decision with good interpretablity.   

\section{Methods}
\subsection{Overall framework}
The overall framework of the proposed evidence-aware multi-modal data fusion is shown in Fig.~\ref{fig1}. In general, it contains backbone classification networks, evidence estimation networks, and the DST fusion module. The backbone network commonly includes two branches~\cite{hoang2022clinically,nguyen2021climat,tolpadi2020deep}, namely, image branch and non-image branch. Concretely, each branch comprises a feature extraction backbone and a classifier to output a probability based on the corresponding data. The features from two branches are concatenated to form the third branch and are attached to a third classifier for prediction.  

\begin{figure}[h]
\begin{center}
\includegraphics[width=\textwidth]{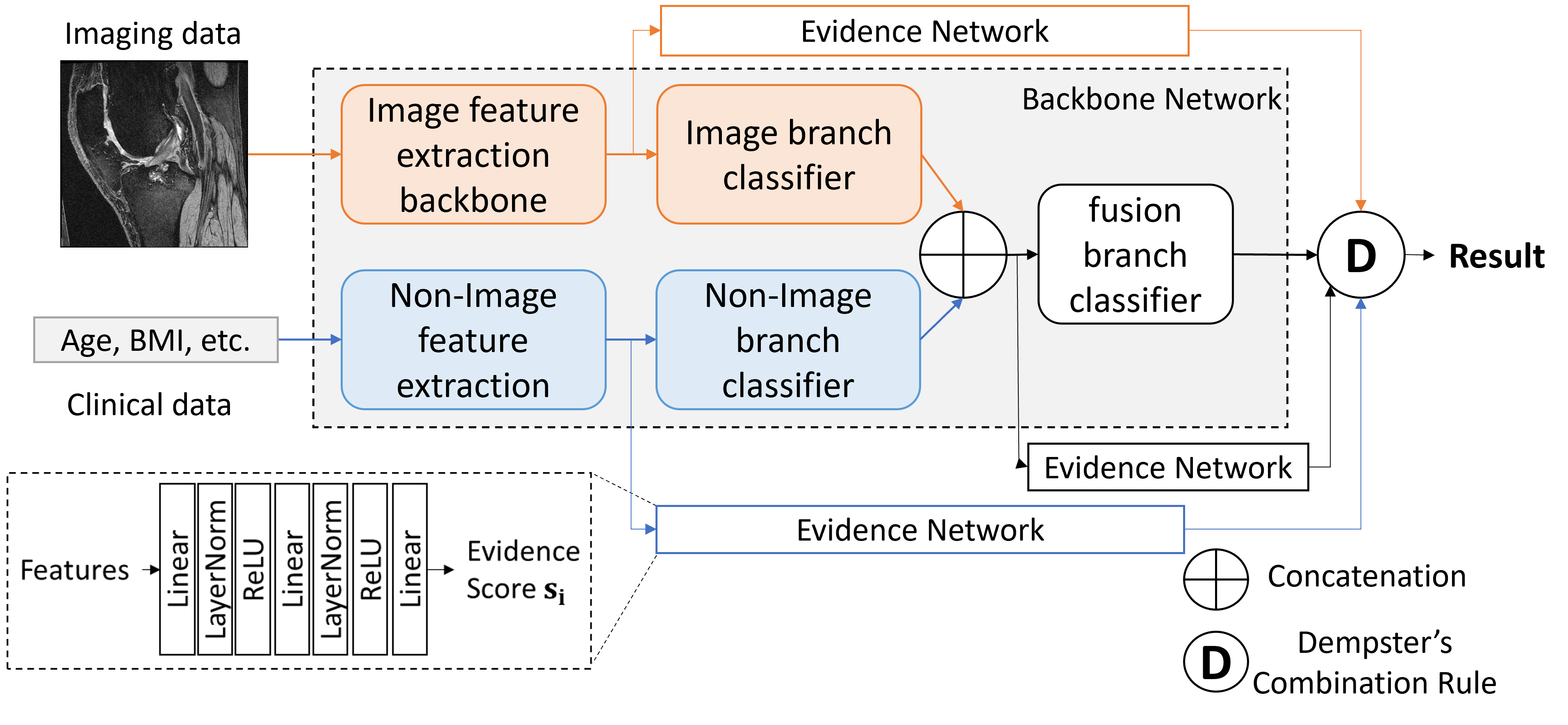}
\caption{Illustration of the evidence-aware multi-modal data fusion framework. It is composed image branch, non-image branch, and the fusion branch. For each branch, there is an evidence network that takes the features as input and outputs the evidence score of the prediction. The evidence score and the probability of each branch are combined based on the Dempster's combination rule.} \label{fig1}
\end{center}
\end{figure}

 Decision-making level fusion of multi-modal data are shown to be effective in medical condition prediction~\cite{hoang2022clinically,nguyen2021climat,tolpadi2020deep}. In our task, considering that the quality of various modalities may vary, the decisions from each branch vary too, which results in conflict of prediction. The aim of our work is to provide a reliable fusion strategy for multi-modal data to improve the prediction accuracy. To this end, we introduce an evidence network to output an evidence score for each branch. The evidence network is designed to estimate the reliability of the data and the output of each branch. Finally, Dempster's combination rule is employed to fuse the information of image branch, non-image branch, and fusion branch according to the estimated evidence scores to obtain the final prediction results. In this way, the decision from each branch can be considered accordingly based on the reliability estimation, alleviating the clash of the decision during the decision fusion stage, leading to a reliable prediction.

\subsection{Evidence network}
\label{evidence_net_section}
To reliably use the decisions from multiple modalities, we need to estimate the reliability of each modality. However, the backbone networks alone are not able to determine whether the data source is of good use and the output is reliable or not. Therefore, an additional network branch, named as evidence network, is employed to evaluate the reliability of the output results. As illustrated in Fig.~\ref{fig1}, we introduce three evidence networks for image branch, non-image branch, and feature fusion branch, respectively. The input of the evidence network is the extracted features of corresponding branch and the output indicates the reliability. The evidence network contains a few linear layers, layer normalisation layers, and non-linear ReLU activation functions.  

We define the reliability of the current output $p_i$ of branch $i$ as evidence score $s_i \in [0,1]$. The more reliability a decision has, the larger its $s_i$ is. This means the current branch is making a confident and right decision. On the contrary, when the branch makes a wrong decision, the $s_i$ needs to be a small value, indicating the output is of high uncertainty and not reliable. Given a well-trained classifier of branch $i$, we pass all the samples $x^{j}|_{j=1}^{N}$, where $N$ is the number of samples, through the network for classification and record if it makes a mistake or not. When classifier of the $i^{th}$ branch returns a wrong decision for a sample $x^{j}$, this means the decision $p_i^{j}$ are not reliable, and the $s_i^{j}$ for this sample should be $0$. On the contrary, when the classifier of branch $i$ returns an accurate result for the sample, $s_i^{j}$ should be 1, denoting a trust-worthy decision. We take these binary scores as the evidence network's learning target $s_i^{j*}$. The evidence networks are trained to generate the evidence scores $\hat{\mathbf{s}}$ that are close to the true evidence score $\mathbf{s^{*}}$. 

\subsection{Dempster's combination rule}
Dempster-Shafer theory (DST)~\cite{dempster1967upper,dempster2008upper}, also known as evidence theory or the theory of belief functions, is a framework for reasoning with uncertainty. DST combines the evidence from various sources with a degree of belief. Studies have shown DST offers superior fusion results in many field~\cite{denoeux2000neural,denoeux2008k,huang2022evidence,huang2021covid,li2022confidence}. The Dempster's combination rule is based on DST and is applied to reasonably fuse the estimated evidence scores $\hat{\mathbf{s}}$ with the probability $\mathbf{p}$ for all the branches. 

Specifically, let $\Omega=\{\omega_1, \omega_2, ..., \omega_K\}$ be a list of $K$ hypotheses, and the evidence can be represented by a basic probability assignment, which is a mapping $m(\cdot): 2^{\Omega} \rightarrow[0,1]$. We have $\sum_{A\subseteq \Omega}{m(A)=1}$, and $m(\cdot)$ is called a mass function. 

The evidence of two mass function $m_1(X_1)$ and $m_2(X_2)$ can be fused by the Dempster's combination rule:
\begin{equation}
    m_f(A) = \frac{1}{M} \sum_{{X_1} \cap {X_2} = A} ~{m_1({X_1})}{m_2({X_2})},
\label{eq_fuse}    
\end{equation}
where $M=\sum_{{X_1} \cap {X_2} \neq \emptyset} {m_1({X_1})}{m_2({X_2})}$. Eq.~(\ref{eq_fuse}) is extendable to multiple evidence:  

\begin{equation}
    m_f(A) = \frac{1}{M} \sum_{{X_1} \cap {X_2} \cap ... {X_T}= A} ~{\prod_{i=1}^{T}m_i({X_i})},  
    \label{eq_fuse_multi}  
\end{equation}
where $M=\sum_{{X_1} \cap {X_2} \cap ... {X_T} \neq \emptyset} \prod_{i=1}^{T}m_i({X_i})$.

In our task, we have two hypotheses, positive and negative, so we define $\Omega = \{T, F\}$. $T$ represents the positive outcome, and $F$ represents the negative outcome. Then, we can construct the set $\{\varnothing, T, F, U\}$, where $U = \{T, F\}$ represents the uncertain prediction. 

The mass function for each branch $i$ can be defined as the evidence score calibrated output. We ignore the sample index $j$ here for simplicity:
\begin{equation}
 \begin{aligned}
 \begin{cases}
m_i(T) =  \hat{s_i} p_i; \\
m_i(F) = \hat{s_i} (1-p_i);  \\ 
m_i(U) = 1 -  \hat{s_i}.
 \end{cases}
 \end{aligned}
\end{equation}

We take two pieces of evidence as an example, the combination of two branches can be formulated as:
\begin{flalign}
& m_f(T) =  \frac{1}{M} [m_1(T)m_2(T) + m_1(T)m_2(U) + m_2(T)m_1(U)];\\
& m_f(F) = \frac{1}{M} [m_1(F)m_2(F) + m_1(F)m_2(U) + m_2(F)m_1(U)];\\ 
& m_f(U) =  \frac{1}{M} m_1(U)m_2(U),   
\label{fuse_mass}
\end{flalign}
where $M=m_f(T)+m_f(F)+m_f(U)$ is the normalisation factor. The combination rule can be extended to multiple branches according to equation (\ref{eq_fuse_multi}). 

\subsection{Network training}
We first train the backbone classification networks with the cross entropy loss. Once the backbone network is fully-trained, we fixed the weights and generate the evidence score labels as described in section \ref{evidence_net_section}. For each branch, we train an evidence network with a mean absolute error (MAE) loss $L_{evid} = \frac{1}{N} \sum_{j=1}^{N} |\hat{\mathbf{s}}^j - \mathbf{s}^{j*}|$. The true evidence scores are imbalanced in the training samples, where the majorities are 1, and the minorities are 0. Therefore, similar to \cite{tolpadi2020deep}, we duplicated the minority cases during the training to make the dataset balanced. 

\section{Experiments and results}
\subsection{Datasets}
In this study, we apply the proposed methods to TKR prediction and follow the settings in \cite{tolpadi2020deep} to conduct the experiments. The data was obtained from Osteoarthritis Initiative (OAI) \cite{peterfy2008osteoarthritis} dataset. 3D Double Echo Steady-state (DESS) MRI images are used as the imaging data, and 27 non-imaging variables are used for the non-image predictions, as in \cite{tolpadi2020deep}.

The DESS MRIs were cropped in the center region to have size of $320 \times 320 \times 120$. The patients who underwent TKR in 5 years are labelled as positive cases; otherwise, the samples are non-TKR control group. The class imbalance problem is significant in the dataset as the majority of the patients did not get the knee replaced. To avoid the bias introduced by the class imbalance, which can affect the evaluation of the model \cite{zhou2021review}, we randomly sample the non-TKR cases to have a similar number of TKR cases in our dataset. We repeated the random sampling process to create three datasets to evaluate the proposed method. In total, there are 1,717 samples in each of our dataset, and we use $80 \%$ for training, $10 \%$ for validation, $10 \%$ for testing.    

\subsection{Network configuration and experimental setups}
The experiments were conducted on an NVIDIA Tesla V100 GPU and we implemented the networks with PyTorch library. The evidence networks in our study contain three interleaved linear, normalisation and ReLU layers, and the size of all the linear layers is 32. We train the evident networks for 100 epochs and saved the best model on the validation dataset. We use the Adam optimiser with a learning rate of 0.0001 and a weight decay to prevent over-fitting.   

There are three settings in our experiments, which are image-only, clinical-only, and multi-modal on the three datasets. The backbone networks are mainly based on~\cite{tolpadi2020deep} and~\cite{hoang2022clinically,nguyen2021climat}. For image-only setting, we employed DenseNet-121 as our backbone using the code and configurations from~\cite{tolpadi2020deep}.  For the clinical-only experiments,  we trained three models. (1) logistic regression (LR): we first used a simple LR method to perform the classification \cite{tolpadi2020deep}. (2) fully-connected network (FC): since LR is a simple approach and only contain one linear layer, we improved the non-imaging branch with a fully-connected network (FC). In FC, the categorical variables go through embedding layers, and the continuous variables are fed into linear layers. After layer normalisation and ReLU activation function, the projected features are summed up and fed into another linear layer. (3) Transformer (Trans): to further improve the non-imaging branch, we use the transformer architecture \cite{hoang2022clinically,nguyen2021climat}. Then, we implemented multi-modality models and our proposed method. The image model is combined with each of the three clinical models, respectively. The LR in multi-modality section of Table~\ref{tabs_main} combined imaging data with non-imaging data with logistic regression (LR) for multi-modal prediction~\cite{tolpadi2020deep}. For FC and Trans in the multi-modality part of Table~\ref{tabs_main}, the corresponding imaging and non-imaging branches' features are concatenated and fed into a third fusion branch, which contains a linear layer for the decision making. Finally, we implemented the proposed fusion method on all baselines, labelled as DST in Table~\ref{tabs_main}. Specifically, the features from the backbone networks are used as the evidence network input. The input of the evidence network for imaging branch draws from the DenseNet-121 and the feature size is 1018. The input of the evidence network for LR, FC, Trans and fusion branch has feature size of 34, 32, 64, and 4, respectively. We use accuracy, specificity, sensitivity, precision, F1 score and area under the ROC curve (AOC) in the scikit-learn package to evaluate the algorithms' performance. 

\subsection{Algorithms comparison}
The comparison of various methods over all datasets is shown in Table~\ref{tabs_main}. The image-only predictions have good accuracy, which are around 82$\%$, 81$\%$ and 83$\%$ on datasets 1, 2,and 3, respectively. In comparison, the clinical-only predictions are normally inferior, which are less than 80$\%$, except that FC and Trans on dataset 3 are close to 85$\%$.

\begin{table}[h]
\centering
\caption{Performance comparison of different methods on OAI dataset. The bold number are the improved results.}\label{tabs_main}
\begin{tabular}{c l l r r r r r r}
\hline
\hline
\textbf{Data} & \multicolumn{2}{c}{\textbf{Method}} & \textbf{Acc.} & \textbf{Spec.} & \textbf{Sens.} & \textbf{Prec.} & \textbf{F1} & \textbf{AUC}  \\ \hline
 \hline
\multirow{9}*{\textbf{1}} & \multicolumn{2}{l}{\textbf{Image-only}} & 82.08 & 86.00 & 76.71 & 71.20 & 78.32 & 81.36   \\ \cline{2-9}
& \multirow{3}{*}{\textbf{Clinical-only}} & \textbf{LR} & 72.25 & 92.00 & 45.21 & 59.51 & 57.89 & 68.60   \\
& & \textbf{FC} & 79.77 & 83.00 & 75.34 & 67.96 & 75.86 & 79.17 \\
& & \textbf{Trans} & 78.61 & 82.00 & 73.97 & 66.46 & 74.48 & 77.99 \\ \cline{2-9}
& \multirow{6}{*}{\textbf{Multi-modal}}& \textbf{LR} & 84.39 & 86.00 & 82.19 & 74.16 & 81.63 & 84.10\\
& & \textbf{LR (DST)}& \textbf{85.55} & \textbf{89.00} & 80.82 & \textbf{76.21} & \textbf{82.52} & \textbf{84.91}\\ \cline{3-9}
& & \textbf{FC} & 84.39 & 89.00 & 78.08 & 74.70 & 80.85 & 83.54\\
& & \textbf{FC (DST)} & \textbf{85.55} & \textbf{91.00} & \textbf{78.08} & \textbf{76.68} & \textbf{82.01} & 
\textbf{85.55} \\ \cline{3-9}
& & \textbf{Trans} & 83.82 & 93.00 & 71.23 & 74.92 & 78.79 & 82.12\\
& & \textbf{Trans (DST)} & \textbf{84.39} & 92.00 & \textbf{73.97} & \textbf{75.41} & \textbf{80.00} & \textbf{82.99} \\  
\hline \hline

\multirow{9}*{\textbf{2}} & \multicolumn{2}{l}{\textbf{Image-only}} & 80.92 & 85.00 & 75.34 & 69.60 & 76.92 & 80.17   \\ \cline{2-9}
& \multirow{3}{*}{\textbf{Clinical only}} & \textbf{LR} & 69.94 & 88.00 & 45.21 & 56.27 & 55.93 & 66.60   \\
& & \textbf{FC} & 77.46 & 84.00 & 68.49 & 65.18 & 71.94 & 76.25 \\
& & \textbf{Trans} & 80.35 & 84.00 & 75.34 & 68.77 & 76.39 & 79.67 \\ \cline{2-9}
& \multirow{6}{*}{\textbf{Multi-modal}}& \textbf{LR} & 83.81 & 87.00 & 79.45 & 75.38 & 80.56 & 83.23\\
& & \textbf{LR (DST)}& \textbf{86.71} & \textbf{91.00} & \textbf{80.82} & \textbf{78.22} & \textbf{83.69} & \textbf{85.91}\\ \cline{3-9}
& & \textbf{FC} & 83.82 & 86.00 & 80.82 & 73.41 & 80.82 & 83.41\\
& & \textbf{FC (DST)} & \textbf{84.97} & 85.00 & \textbf{84.93} & \textbf{74.74} & \textbf{82.67} & \textbf{84.97} \\ \cline{3-9}
& & \textbf{Trans} & 84.97 & 89.00 & 79.45 & 75.46 & 81.69 & 84.23\\
& & \textbf{Trans (DST)} & \textbf{85.55} & \textbf{89.00} & \textbf{80.82} & \textbf{76.01} & \textbf{82.76} & \textbf{85.10} \\  
\hline \hline

\multirow{9}*{\textbf{3}} & \multicolumn{2}{l}{\textbf{Image-only}} & 83.24 & 87.00 & 78.08 & 72.83 & 79.72 & 82.54   \\ \cline{2-9}
& \multirow{3}{*}{\textbf{Clinincal only}} & \textbf{LR} & 80.35 & 94.00 & 61.64 & 70.58 & 72.58 & 77.82   \\
& & \textbf{FC} & 84.97 & 86.00 & 83.56 & 74.90 & 82.43 & 84.78 \\
& & \textbf{Trans} & 85.55 & 87.00 & 83.56 & 75.82 & 82.99 & 85.28 \\ \cline{2-9}
& \multirow{6}{*}{\textbf{Multi-modal}}& \textbf{LR} & 87.86 & 88.00 & 87.67 & 79.03 & 85.91 & 87.84\\
& & \textbf{LR (DST)}& \textbf{90.75} & \textbf{92.00} & \textbf{89.04} & \textbf{83.91} & \textbf{89.04} & \textbf{90.52}\\ \cline{3-9}
& & \textbf{FC} & 89.02 & 90.00 & 87.67 & 81.03 & 87.07 & 88.84\\
& & \textbf{FC (DST)} & \textbf{90.17} & \textbf{92.00} & \textbf{87.67} & \textbf{83.13} & \textbf{88.28} & \textbf{89.84} \\ \cline{3-9}
& & \textbf{Trans} & 89.02 & 90.00 & 87.67 & 81.03 & 87.07 & 88.84\\
& & \textbf{Trans (DST)} & \textbf{89.60} & \textbf{92.00} & 86.30 & \textbf{82.36} & \textbf{87.50} & \textbf{89.15} \\  
\hline \hline
\end{tabular}
\end{table}
 The multi-modal networks without DST combine the information from both sources and consistently obtain better results than the single-model predictions on all datasets. The improvement in accuracy of the multi-model networks are around 2-4$\%$ higher than the best single model. This means that combination of information from multiple sources improves the prediction performance. 

However, the multi-modal networks without DST still conduct a simple combination treating all the data sources and branches equally, and cannot make good prediction when multiple branches output conflicting predictions and have significantly different reliability. The proposed evidence-aware approach, labelled as DST, considers the reliability of the source data and the network prediction, greatly reducing the uncertainties of the prediction. As illustrated in Table~\ref{tabs_main}, the methods with DST almost consistently outperform the counterparts in terms of all metrics on all datasets. This shows the effectiveness of the proposed evidence network and fusion strategy.

\subsection{Additional analysis}
\label{analysis_section}
We further investigated the effectiveness of the proposed methods. First, we compared the proposed fusion strategy with the average fusion method \cite{li2022confidence}, which simply averages the classification probabilities of all the branches. The comparison results of the proposed method (DST) and the average fusion (Avg.) is shown in Table~\ref{tabs_fuse}. The proposed methods have around 1$\%$ higher accuracy compared to the average fusion method on dataset 1. The improvement is also seen on datasets 2 and 3. 
Therefore, it is important to consider the reliability of each branch's output for final results.

\begin{table}[h!]
\centering
\caption{Prediction accuracy between the proposed method (DST) and the average fusion (Avg.)~\cite{li2022confidence} on all dataset with different baseline networks. D1, D2, D3 represent dataset 1, 2, 3, respectively.}\label{tabs_fuse}
\begin{tabular}{l|c|c|c|c|c|c|c|c|c}
\hline
Acc. & D1 LR & D1 FC & D1 Trans & D2 LR & D2 FC & D2 Trans & D3 LR & D3 FC & D3 Trans \\
\hline
DST & 85.55 & 85.55 & 84.39 & 86.71 & 84.97 & 85.55 & 90.75 &  90.17 &  89.60\\ 
\hline
Avg. & 84.60 & 84.97 & 83.24 & 83.24 & 84.97 & 85.55 & 87.86 & 89.60 & 89.02\\ \hline
\end{tabular}
\end{table}

Second, we examine the evidence score estimation to confirm that the evidence network has indeed learned to perform its task. Specifically, we check the evidence scores of correctly classified samples and misclassified samples, respectively. The correctly classified samples should have larger evidence scores, and the evidence scores of the rest samples should be near $0$. We use histogram to visualise it. Fig.\ref{figs1}-\ref{figs4} in the supplementary shows an example histogram of the fuse branch of LR backbone on dataset 1. For most of the correct results, the scores falls near $1$; and for the wrong classifications, the scores are much smaller and close to $0$. This means the evidence network can learn the reliability of the backbone networks. 

\section{Discussion and conclusion}
In this paper, we propose a novel evidence-aware multi-modal data fusion strategy for medical condition prediction based on DST. The proposed method considers the reliability of the source data and the output of each modality. We apply our methods to TKR prediction task and the experiments show the increased prediction accuracy of the proposed approach. Although the experiments are conducted on DESS MRI and clinical data from OAI dataset, other source of data, such as bio-mechanical analysis, can also be included by adding an additional branch with the corresponding evidence network. The proposed approach is also applicable to other medical condition prediction tasks that require the information from multiple modalities.

\bibliographystyle{splncs04}
\bibliography{mybibliography}

\newpage
\section{Supplementary Materials}
\begin{figure}[h!]
\setcounter{figure}{0}
\renewcommand{\thefigure}{S\arabic{figure}}
\begin{center}
\includegraphics[width=\textwidth]{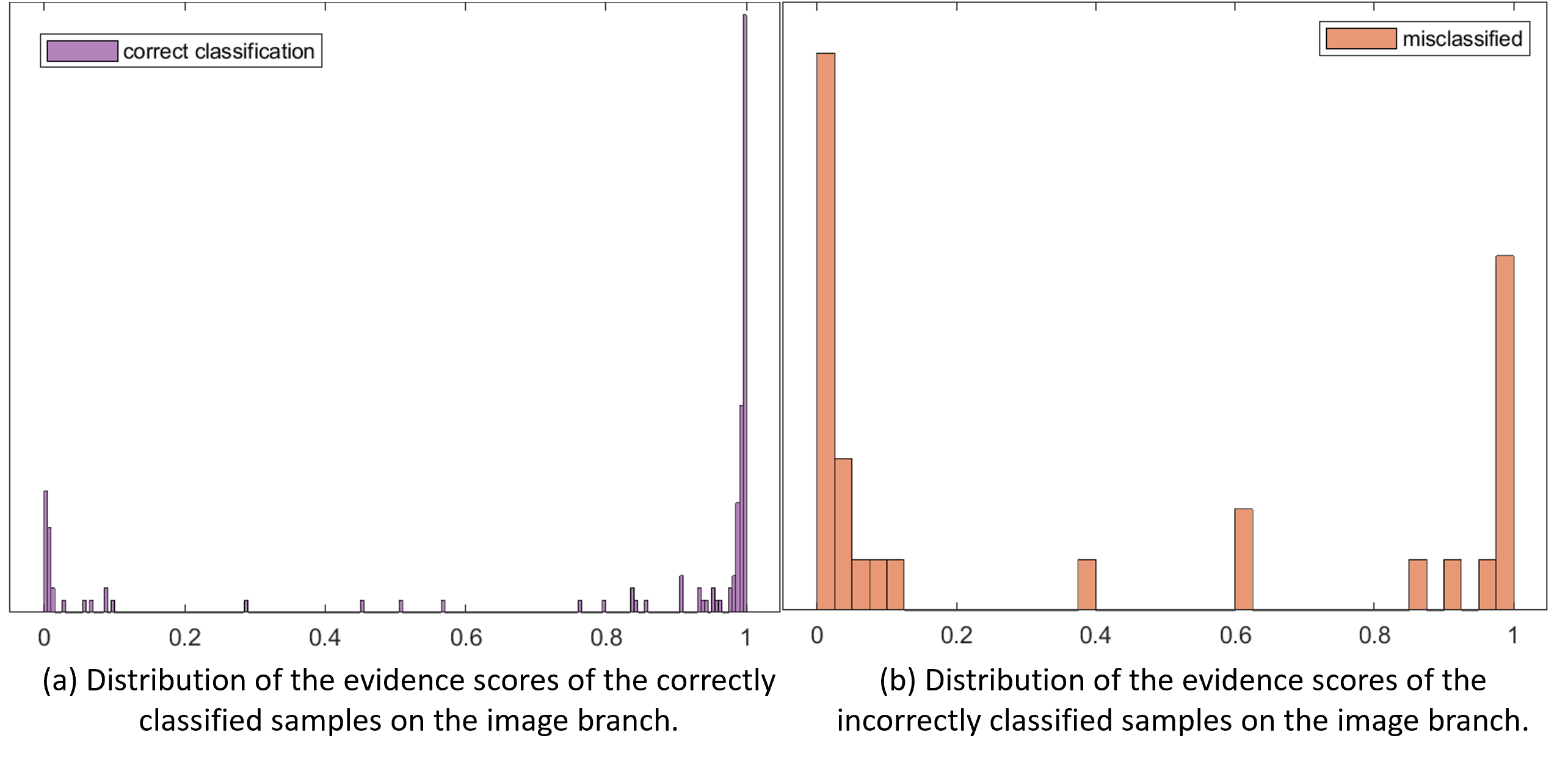}
\caption{Histogram of the evidence scores for the image branch of the LR (DST) model on dataset 1.} \label{figs1}
\end{center}
\end{figure}

\begin{figure}[h!]
\renewcommand{\thefigure}{S\arabic{figure}}
\begin{center}
\includegraphics[width=\textwidth]{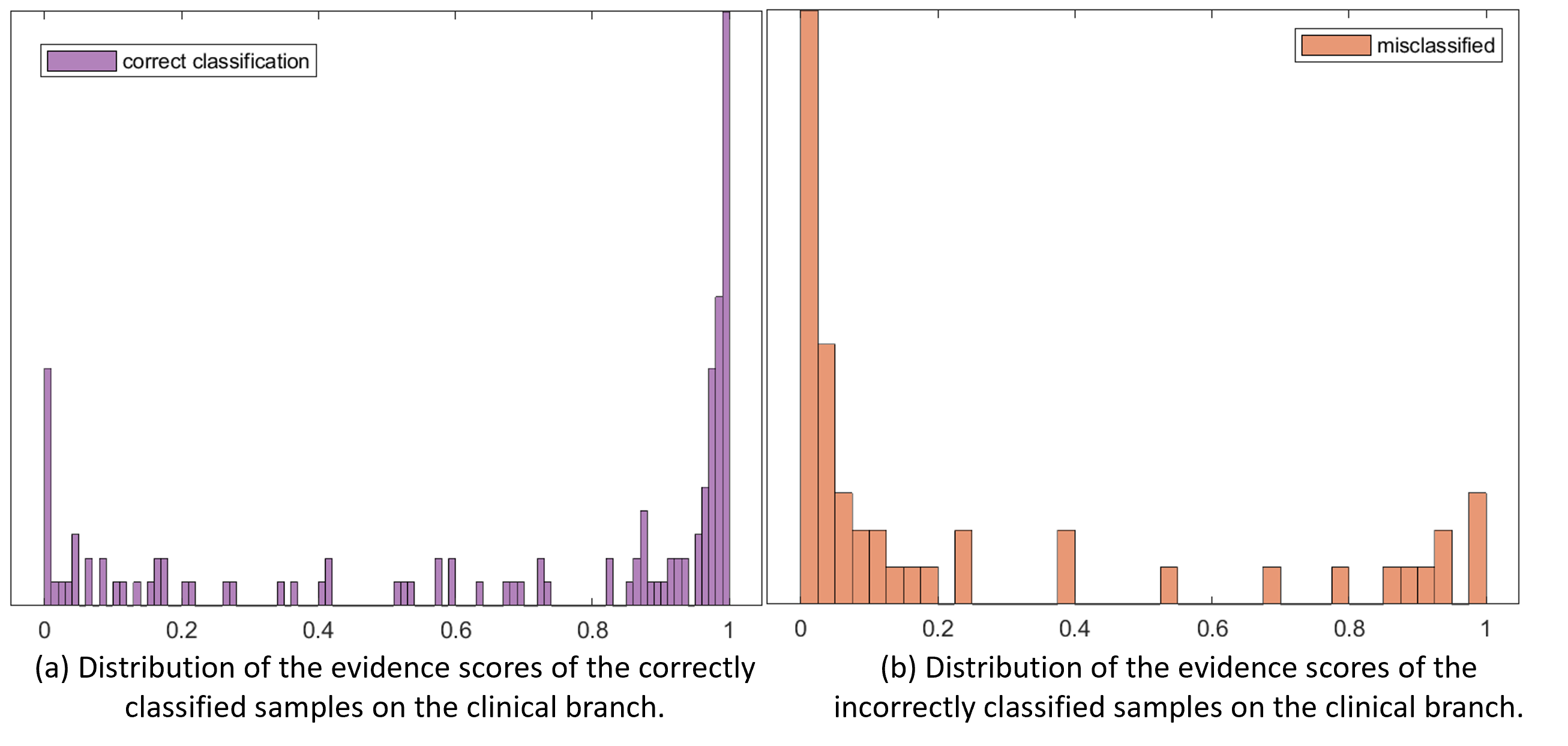}
\caption{Histogram of the evidence scores for the clinical branch of the LR (DST) model on dataset 1.}
\label{figs2}
\end{center}
\end{figure}

\begin{figure}[h!]
\renewcommand{\thefigure}{S\arabic{figure}}
\begin{center}
\includegraphics[width=0.6\textwidth]{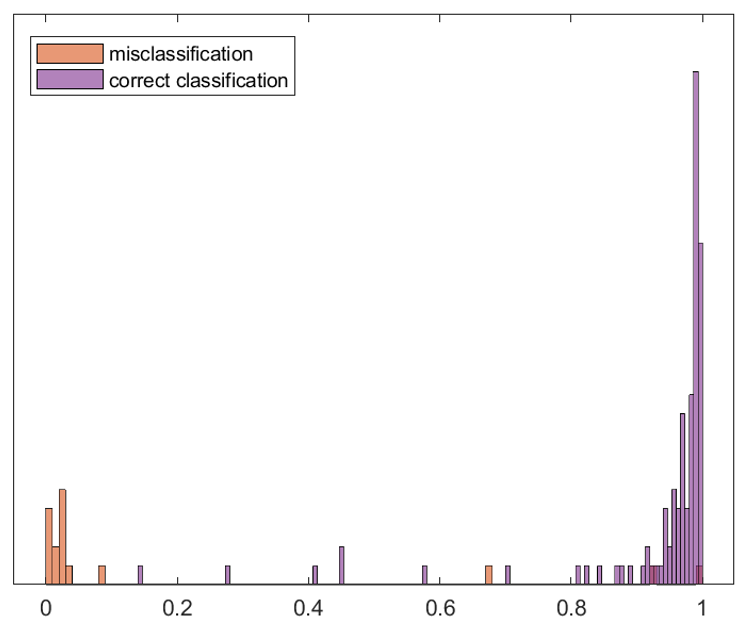}
\caption{Histogram of the evidence scores for the fusion branch of the LR (DST) model on dataset 1. The orange is the scores for the misclassified samples, and the purple is the scores for the correct samples.}
\label{figs3}
\end{center}
\end{figure}

\begin{figure}[h!]
\renewcommand{\thefigure}{S\arabic{figure}}
\begin{center}
\includegraphics[width=0.6\textwidth]{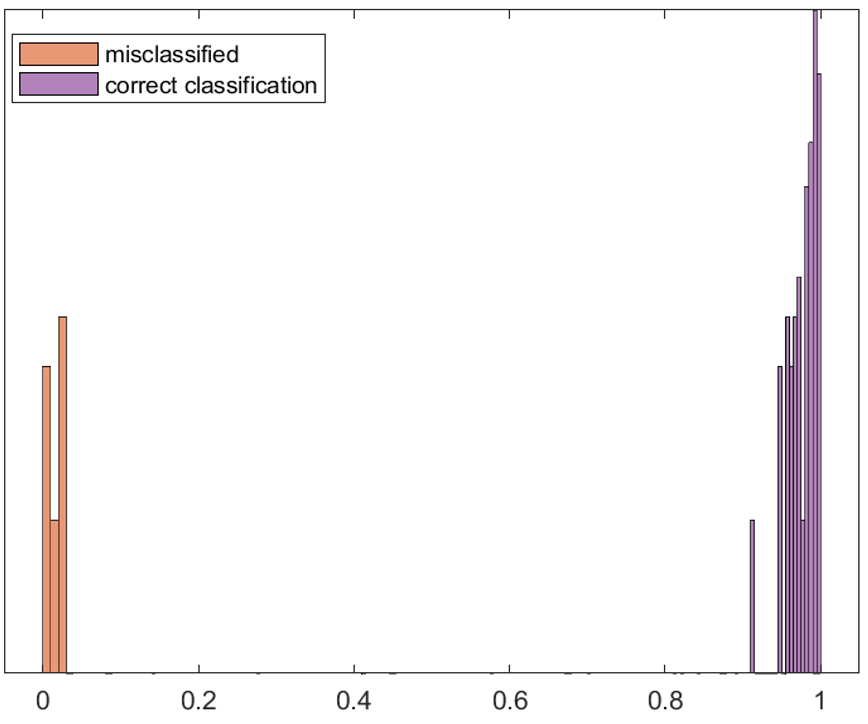}
\caption{We collect all the correct samples from three branches of the LR (DST) model on dataset 1 and visualise the evidence scores in purple. The histogram in orange is the evidence score distribution of the misclassified samples of all branches}
\label{figs4}
\end{center}
\end{figure}

\end{document}